\documentclass[letterpaper]{article} 
\usepackage{aaai2026}  
\nocopyright  
\usepackage{times}  
\usepackage{helvet}  
\usepackage{courier}  
\usepackage[hyphens]{url}  
\usepackage{graphicx} 
\usepackage{natbib}  
\usepackage{caption} 
\usepackage{algorithm}
\usepackage{algorithmic}

\usepackage{booktabs}
\usepackage{pifont}
\usepackage{multirow}
\usepackage{amsmath}
\usepackage{amssymb}

\usepackage{newfloat}
\usepackage{listings}
\DeclareCaptionStyle{ruled}{labelfont=normalfont,labelsep=colon,strut=off} 
\floatstyle{ruled}
\newfloat{listing}{tb}{lst}{}
\floatname{listing}{Listing}

\title{Wavefront-Constrained Passive Obscured Object Detection}

\author{
    Zhiwen Zheng\textsuperscript{\rm 1,\rm 2}\thanks{Equal Contribution.}
    Yiwei Ouyang\textsuperscript{\rm 1,\rm 2}\footnotemark[1]
    Zhao Huang\textsuperscript{\rm 3},
    Tao Zhang\textsuperscript{\rm 1},
    Xiaoshuai Zhang\textsuperscript{\rm 4}\thanks{Corresponding Authors.}
    Huiyu Zhou\textsuperscript{\rm 5},
    Wenwen Tang\textsuperscript{\rm 6}\footnotemark[2],
    Shaowei Jiang\textsuperscript{\rm 1}\footnotemark[2],
    Jin Liu\textsuperscript{\rm 1,\rm 2}\footnotemark[2],
    Xingru Huang\textsuperscript{\rm 1,\rm 2}\footnotemark[2]
}

\affiliations{
    \textsuperscript{\rm 1}School of Communication Engineering, Hangzhou Dianzi University, Hangzhou 310018, China\\
    \textsuperscript{\rm 2}Weidian Corporation, Beijing, 100015, China\\
    \textsuperscript{\rm 3}Department of Computer and Information Science, Northumbria University, NE1 8ST Newcastle upon Tyne, UK\\
    \textsuperscript{\rm 4}Faculty of Information Science and Engineering, Ocean University of China, 266404, Qingdao, China\\
    \textsuperscript{\rm 5}School of Computing and Mathematical Sciences, University of Leicester, Leicester LE1 7RH, UK\\
    \textsuperscript{\rm 6}Johns Hopkins University, Baltimore, MD, 21218, USA\\[3pt]

    zhiwen.zheng@hdu.edu.cn, yiwei.ouyang@hdu.edu.cn, zhao.huang@northumbria.ac.uk, tzhang@hdu.edu.cn, x.zhang@ouc.edu.cn, hz143@leicester.ac.uk, wtang18@jhu.edu, jiangsw@hdu.edu.cn, jinliu@hdu.edu.cn, xingru.huang@hdu.edu.cn

}

\begin{document}

\maketitle

\begin{abstract}
Accurately localizing and segmenting obscured objects from faint light patterns beyond the field of view is highly challenging due to multiple scattering and medium-induced perturbations. Most existing methods, based on real-valued modeling or local convolutional operations, are inadequate for capturing the underlying physics of coherent light propagation. Moreover, under low signal-to-noise conditions, these methods often converge to non-physical solutions, severely compromising the stability and reliability of the observation. To address these challenges, we propose a novel physics-driven Wavefront Propagating Compensation Network (WavePCNet) to simulate wavefront propagation and enhance the perception of obscured objects. This WavePCNet integrates the Tri-Phase Wavefront Complex-Propagation Reprojection (TriWCP) to incorporate complex amplitude transfer operators to precisely constrain coherent propagation behavior, along with a momentum memory mechanism to effectively suppress the accumulation of perturbations. Additionally, a High-frequency Cross-layer Compensation Enhancement is introduced to construct frequency-selective pathways with multi-scale receptive fields and dynamically models structural consistency across layers, further boosting the model’s robustness and interpretability under complex environmental conditions. Extensive experiments conducted on four physically collected datasets demonstrate that WavePCNet consistently outperforms state-of-the-art methods across both accuracy and robustness.
\end{abstract}

\begin{links}
    \link{Code}{https://github.com/IMOP-lab/WavePCNet-Pytorch}
\end{links}

\section{Introduction}

\begin{figure}[ht]
    \centering
    \includegraphics[width=\linewidth]{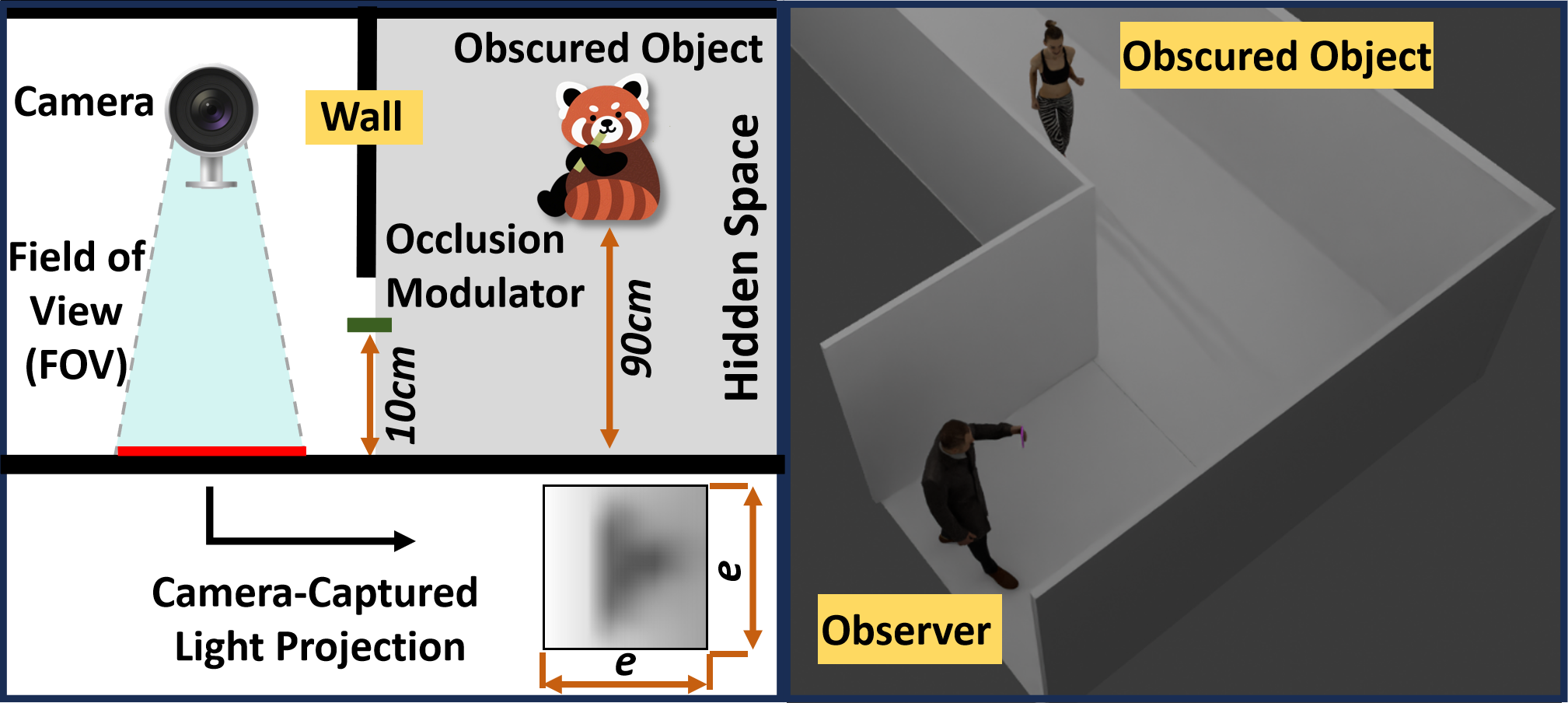}
    \caption{Typical scenario underlying the passive OOD regime, in which the target, entirely occluded by an opaque intervening barrier that prevents any direct radiative conveyance to the sensor, is inferred solely through the diffuse, low-intensity speckle distribution scattered onto the visible relay surface.}
    \label{fig:ChangJing}
\end{figure}

The Non-line-of-sight (NLOS) targets detection has broad real-world applications, including blind-spot collision avoidance in autonomous vehicles, security surveillance through occluding structures, and post-disaster search and rescue scenarios, where these targets are often hidden around corners or fully obscured by opaque barriers \cite{zhao2023intelligent}. These applications pose a challenging question "Is it possible to localize a target hidden behind a wall solely by relying on the faint speckle field generated by ambient light scattering, without any active illumination? "

Accordingly, the work formulates the Obscured Object Detection (OOD) task: from a measured surface-intensity distribution $I_w(\mathbf{x})$ of exceptionally low signal-to-noise ratio, determine both a semantic mask $M(\mathbf{x})$ and a geometric estimate $\mathbf{p}$ of the out of view object. This passive regimen circumvents spectral-allocation and regulatory burdens imposed by emitters, permits direct adoption of commodity imaging assemblies, and thus promises economical deployment \cite{beckus2019multi} \cite{boger2019passive}. Figure \ref{fig:ChangJing} depicts the canonical arrangement: an object entirely eclipsed by an opaque screen, no direct photons admitted to the sensor; yet, from the residual speckle upon the observable wall, the observer infers its position and extent.

Although OOD affords emitter free operation, cost efficiency and deployment simplicity, it exacts inherently feeble observations and entails convoluted physics: multiple reflections and medium roughness rapidly decorrelate phase along propagation paths, impose spectral bandlimitation and deplete high-frequency content, such that only low-SNR, phase mismatched speckle appears on accessible surfaces \cite{khawaja2019survey}. Models built on real-valued convolution kernels cannot represent the phase perturbations and path-integral effects intrinsic to coherent scattering \cite{rotter2017light}. Hence two principal obstacles arise: first, in occluded environments, convoluted propagation trajectories and severe wavefront distortion may drive the inversion to settle in unphysical or multi-valued solution spaces \cite{may2021fast};   second, indirect reflection induces energy decay and high frequency detail loss, manifesting as blurred edges and fractured structural features \cite{cao2022shaping}.

To overcome these challenges, we propose WavePCNet, a novel framework tailored for inferring obscured objects from a single speckle field, which consists of Tri-Phase Wavefront Complex-Propagation Reprojection (TriWCP) and High frequency Cross-layer Compensation. The TriWCP, grounded in Fresnel diffraction, forges a three-stage complex-domain optimisation pathway; a momentum-retentive update confines the trajectory to a physically admissible low-rank sub-space, curbing multiplicity driven by wavefront distortion. The Cross-layer Compensation, via multi-scale frequency-selective filtering, amplifies residual high-frequency cues, while a semantics-guided attention stream aligns structural detail across layers, thus conjointly repairing blurred edges and fragmented semantics.

The main contributions are summarized as follow:
\begin{itemize}
\item Introduce WavePCNet, a novel physics-driven framework for passive detection of obscured objects beyond the line of sight. By unifying complex-amplitude propagation with frequency–structure compensation, WavePCNet achieves state-of-the-art accuracy and robustness on our physically collected datasets, even under extreme occlusion and low-SNR conditions.

\item Propose the TriWCP module, which pioneers a three-phase complex-domain optimization strategy grounded in Fresnel diffraction. It incorporates an OTF-consistent complex propagation operator and a momentum memory mechanism to guide the solution within a physically valid, low-rank subspace, effectively mitigating ambiguity from wavefront distortion.

\item Design the Cross-Layer Compensation module, which integrates multi-scale frequency-selective filtering with semantics-guided attention to enhance both high-frequency detail restoration and cross-layer semantic coherence, enabling more precise reconstruction of fragmented object structures.
\end{itemize}

\section{Related Work}

Visible-domain object detection comprises multiple sub-tasks, notably Salient Object Detection (SOD) \cite{wei2024weakpcsod} and Camouflaged Object Detection (COD) \cite{ruan2025mm}.  Models such as U2-Net \cite{QIN2020107404}, PFPN \cite{wang2020progressive}, F3Net \cite{wei2020f3net} represent typical SOD pipelines that adopt multi-scale aggregation and edge refinement, raising localisation fidelity within salient extents; COD, confronting background homogeneity and boundary ambiguity, invokes frequency domain representation, contextual fusion, prompt guidance, as implemented in FAPNet \cite{zhou2022feature}, FDNet \cite{zhong2022detecting}, VSCODE \cite{luo2024vscode}, enlarges representational capacity.  Generative counterparts, CamoDiffusion \cite{chen2024camodiffusion}, Camoformer \cite{yin2024camoformer}, together with branched backbones DC-Net \cite{zhu2025dc}, ZoomNeXt \cite{pang2024zoomnext}, concentrate on camouflage edge modelling and detail amplification;  RefCOD\cite{zhang2025referring} appends linguistic cues, improving multi-instance semantic disentanglement under concealment.  Progress centres on fusion, edge adjustment, spectrum modelling;  unaddressed remains coherent, phase-coupled indirect transport, hence full occlusion outside the field of view escapes current visible-domain formulations.

Non-line-of-sight (NLOS) imaging supplies an alternative trajectory: indirect radiative transport characterised, hidden geometry or radiance inferred.  Transient schemes, e.g., NeTF \cite{shen2021non}, utilise time-of-flight sampling;  coherent formulations, Liu et al. \cite{liu2019non}, inject phase-propagation analysis for strong scattering;  millimetre-wave radar localises obscured objects through penetration \cite{scheiner2020seeing}, infrared thermography retrieves outlines via thermal radiative fields \cite{yuan2024non}.  Structural accuracy emerges, yet specialised hardware, environmental stability, cost burdens persist;  focus settles on physical-field regression, not semantic discrimination.  Hybrid networks embedding physical priors mitigate expense, still confine modelling to signal synthesis, low SNR structural recognition and boundary restoration remaining unresolved.

\section{Method}

\begin{figure*}[ht]
    \centering
    \includegraphics[width=\linewidth]{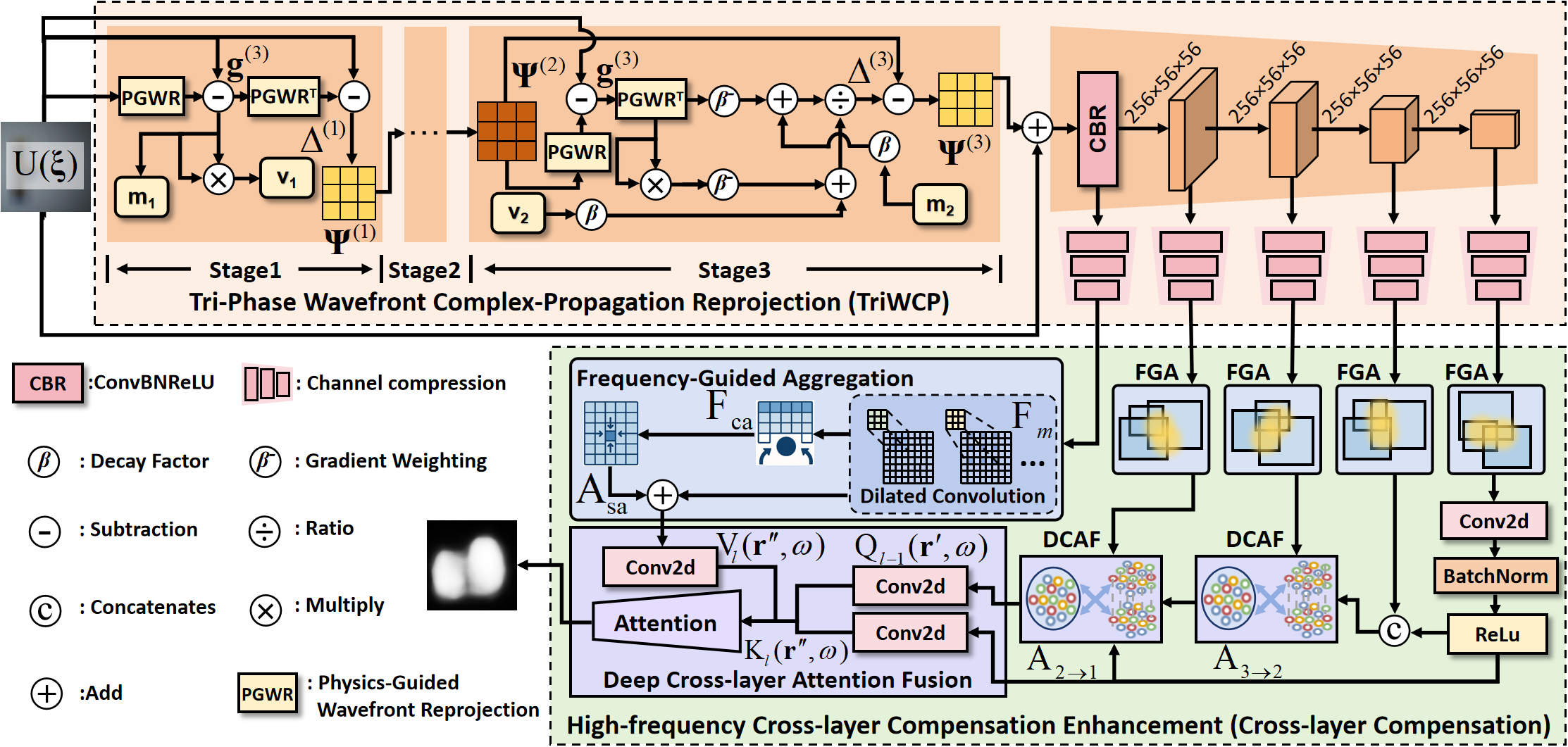}
    \caption{Schematic representation of WavePCNet architecture, wherein a bifurcated topology integrates (i) a physically constrained optimisation pipeline grounded in complex-amplitude propagation (TriWCP), enforcing low-rank spectral admissibility via manifold-constrained trajectory reprojection; and (ii) a cross-layer compensatory stream that augments attenuated high-frequency constituents through semantics-guided, frequency-selective enhancement.}
    \label{fig:all}
\end{figure*}

The study proposed WavePCNet (Figure \ref{fig:all}), a novel physics-driven framework designed to infer obscured objects from a single speckle field in passive, non-line-of-sight scenarios. Specifically, it integrates two key components: (1) Tri-Phase Wavefront Complex-Propagation Reprojection (TriWCP), grounded in Fresnel diffraction, which constructs a three-stage optimization pathway in the complex domain. A momentum-retentive update constrains the solution trajectory within a physically valid low-rank subspace, effectively mitigating ambiguities introduced by wavefront distortion. (2) High-Frequency Cross-Layer Compensation, which enhances residual high-frequency cues via multi-scale frequency-selective filtering, while a semantics-guided attention stream aligns structural details across layers. Together, these components enable robust recovery of blurred edges and fragmented semantics, significantly improving detection stability and accuracy under complex occlusions.

\subsection{Tri-Phase Wavefront Complex-Propagation Reprojection }

\begin{figure*}[ht]
    \centering
    \includegraphics[width=\linewidth]{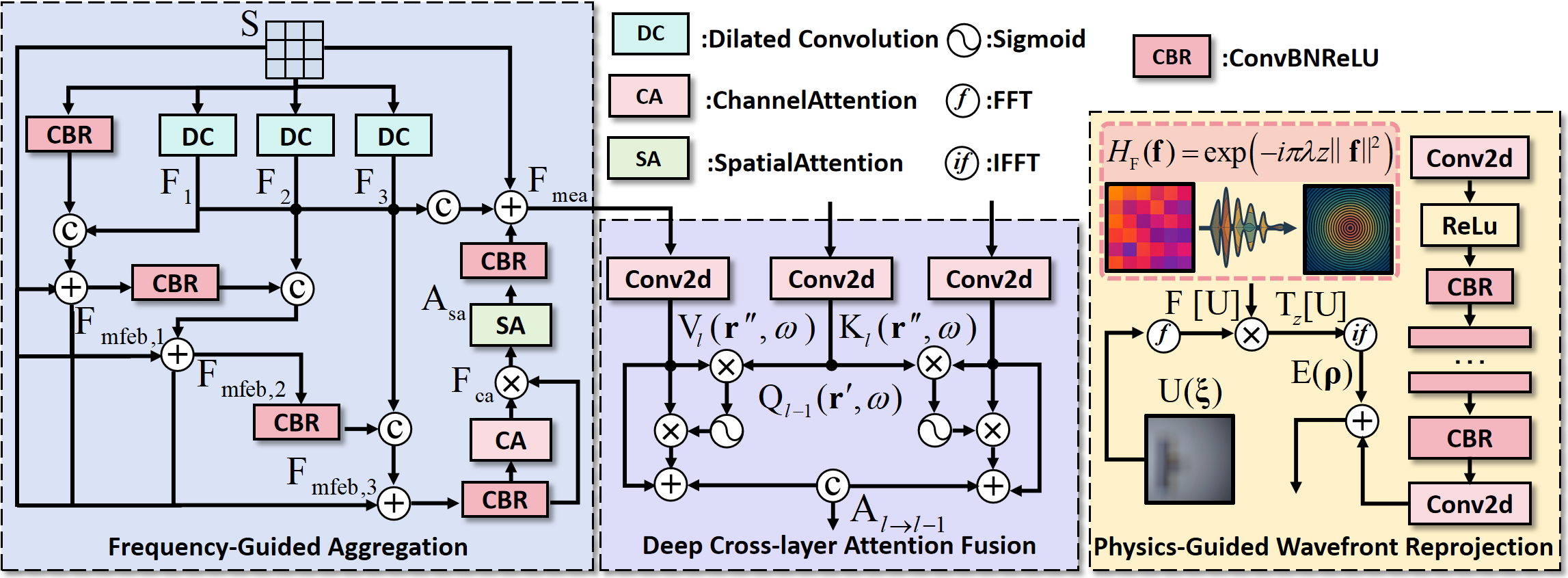}
    \caption{Architectural composition of the proposed modules. Left: Cross-layer Compensation integrates multi-receptive field frequency-selective filtering and semantics-guided structural alignment to recover attenuated high-frequency cues. Center: spatial-frequency attention projection across layers enhances structural consistency. Right: Tri-Phase Complex-Propagation Reprojection constrains optimisation within a physically admissible low-rank subspace via embedded Fresnel transfer operator and memory-augmented updates.}
    \label{fig:model}
\end{figure*}

Propagation of light manifests intrinsic wave behavior and coherence; its advancement adheres to Fresnel diffraction as encapsulated by the complex-amplitude transfer function, thereby confining field evolution to a physically admissible, low-dimensional subspace. To enforce this constraint, we introduce a Tri-Phase Wavefront Complex-Propagation Reprojection module, which comprises a three-stage reconstruction pipeline that explicitly embeds the complex-valued propagation operator, restricts the optimisation trajectory to the coherent low-rank subspace, and reprojects obscured object image structure from the complex wavefront. Such reprojection mitigates unphysical solutions and solution multiplicity. A momentum memory mechanism accumulates integrated responses of historical propagation paths, attenuating high frequency noise and stochastic perturbations. The pipeline is illustrated in Figure \ref{fig:model}.

Under the high frequency scalar approximation, free-space complex amplitude propagation conforms to Fresnel diffraction theory, such that the free propagation process remains strictly governed by the physically permissible propagation operator. Denoting the source plane complex amplitude by $\mathcal{U}(\boldsymbol{\xi})$, the propagation distance by $z$ and the observation plane field by $\mathcal{E}(\boldsymbol{\rho})$, the Hertz-vector wave equation under scalar approximation yields the complex-propagation integral:
\begin{equation}
\mathcal{E}(\boldsymbol{\rho}) = \frac{e^{ikz}}{i\lambda z} \iint_{\mathbb{R}^2} \mathcal{U}(\boldsymbol{\xi}) \exp\left( \frac{ik}{2z} \|\boldsymbol{\rho} - \boldsymbol{\xi}\|^2 \right) \mathrm{d}\boldsymbol{\xi}.
\end{equation}
Let $k=2\pi/\lambda$ denote the wavenumber, i.e., the propagation constant; $\lambda$ the wavelength; $\boldsymbol{\xi},\boldsymbol{\rho}\in\mathbb{R}^2$ the transverse spatial coordinates; the quadratic term $|\boldsymbol{\rho}-\boldsymbol{\xi}|^{2}$ the squared transverse separation, representing the projected path-length difference. The integral thus manifests propagation as a space invariant, convolution type superposition that aggregates global contributions of the source field.Passing to the Fourier domain yields:
\begin{equation}
\begin{split}
\mathcal{E}(\boldsymbol{\rho}) = \mathcal{F}^{-1} \left\{ H_{\text{F}}(\boldsymbol{f}) \cdot \mathcal{F} \left[ \mathcal{U}(\boldsymbol{\xi}) \right] \right\}, \\
H_{\text{F}}(\boldsymbol{f}) = \exp\left( -i\pi \lambda z \|\boldsymbol{f}\|^2 \right),
\end{split}
\end{equation}
where $\mathcal{F}$ and $\mathcal{F}^{-1}$ denote the two-dimensional Fourier transform and its inverse, $\boldsymbol{f}$ the spatial-frequency variable with unit $\mathrm{m}^{-1}$. The multiplicative factor $H_{\mathrm{F}}(\boldsymbol{f})$, known as the complex amplitude transfer function or optical transfer function (OTF), is a pure phase term that effects the Fresnel quadratic-phase modulation of the spectrum.

Under the Fresnel transfer formulation,  free space advancement acts as a complex linear operator that couples a passband frequency filter with quadratic phase modulation. Define, for any $\mathcal{U} \in L^{2}(\mathbb{R}^{2},\mathbb{C})$, the propagation operator $\mathcal{T}_z$ by:
\begin{equation}
\mathcal{T}_z[\mathcal{U}] = \mathcal{F}^{-1} \left[ \chi_{\Omega_z}(\boldsymbol{f}) \cdot \exp\left( -i\pi \lambda z \|\boldsymbol{f}\|^2 \right) \cdot \mathcal{F}[\mathcal{U}] \right],
\end{equation}
where $\chi_{\Omega_z}$ denotes the indicator of the passband
$\Omega_z:={\boldsymbol{f}\in\mathbb{R}^{2}\mid |\boldsymbol{f}|^{2}\le 1/(\lambda z)}$; frequencies outside $\Omega_z$ are suppressed and, within this model, treated as non-propagating.

The physically admissible fields consequently reside in a low-rank complex subspace:
\begin{equation}
\mathbb{S}_z := \left\{ \psi \in L^2(\mathbb{R}^2, \mathbb{C}) \;\middle|\; \psi = \mathcal{F}^{-1} \left[ \chi_{\Omega_z} \cdot \hat{\psi} \right] \right\}.
\end{equation}
In effect, its effective dimensionality scales with the area of $\Omega_z$ and is strictly lower-rank within $L^{2}(\mathbb{R}^{2})$. Any perturbation $\delta\mathcal{U}\notin\mathbb{S}_z$ is removed by propagation. Hence we project search directions onto $\mathbb{S}_z$ via the orthogonal projector:
\begin{equation}
\delta \mathcal{U}_{\text{phys}} = \mathcal{P}_{\mathbb{S}_z}[\delta \mathcal{U}] = \mathcal{T}_z^\dagger \mathcal{T}_z[\delta \mathcal{U}],
\end{equation}
with $\mathcal{P}_{\mathbb{S}_z}$ the orthogonal projection onto $\mathbb{S}_z$ and $\mathcal{T}_z^{\dagger}$ the adjoint of $\mathcal{T}_z$. The discrete Fourier domain derivation appears in Appendix. Constraining updates by $\mathcal{T}_z$ thus restricts the reachable set to $\mathbb{S}_z$, automatically rejecting directions inconsistent with the propagation physics.

In realistic optical systems, the complex field $\Psi^{(z)}(\mathbf{r})$ at any position $\mathbf{r}$ arises from the aggregate superposition of multiple diffraction, interference and scattering trajectories, which collectively form an integral propagation–response operator spanning the full history of propagation.

By introducing a generalized propagator $\mathcal{A}_z^\dagger$, which encapsulates all orders of diffraction, interference and scattering, and by projecting gradients onto the physically admissible subspace $\mathbb{S}_z$, do we prevent non-physical perturbations from accumulating during optimization.  To that end, momentum and energy-spectrum terms are formed via exponential weighting:
\begin{equation}
\boldsymbol{m}^{(t)}(\boldsymbol{r}) := 
\sum_{\tau=1}^{t} \beta_1^{t-\tau} \cdot 
\mathcal{A}_z^\dagger \left[ 
\boldsymbol{g}^{(\tau)}(\boldsymbol{r}) - 
\Pi_{\mathbb{S}_z}[\boldsymbol{g}^{(\tau)}](\boldsymbol{r}) 
\right],
\end{equation}
where the decay factor $\beta_1^{\,t-\tau}$ attenuates early perturbations, imparting an “inertial memory” of the field’s interference structure and thus enhancing convergence stability and directional robustness in complex media.

\begin{equation}
\boldsymbol{v}^{(t)}(\boldsymbol{r}) := 
\sum_{\tau=1}^{t} \beta_2^{t-\tau} \cdot 
\left| 
\mathcal{A}_z^\dagger \left[ 
\boldsymbol{g}^{(\tau)}(\boldsymbol{r}) - 
\Pi_{\mathbb{S}_z}[\boldsymbol{g}^{(\tau)}](\boldsymbol{r}) 
\right]
\right|^2,
\end{equation}
this energy spectrum term capturing the spatial accumulation of perturbations along propagation paths, modulates momentum amplitude and suppresses unstable propagation modes via square-root scaling.  Complete derivation in Appendix.

To attenuate residual high spatial frequency perturbations while maintaining the spatial continuity mandated by physical propagation, a convolutional smoothing corrector is inserted into the momentum update: the previous-step residual
$\boldsymbol{\delta}^{(t-1)}(\boldsymbol{r})=\boldsymbol{g}^{(t-1)}(\boldsymbol{r})-\Pi_{\mathbb{S}_z}[\boldsymbol{g}^{(t-1)}](\boldsymbol{r})$
passes through a low-pass kernel $\mathcal{K}(\boldsymbol{r}-\boldsymbol{r}')$, yielding a locally averaged disturbance which biases the momentum direction away from non-physical oscillatory components and, hence, prevents their admission into the update trajectory. The update at iteration $t$ is defined as:
\begin{equation}
\Delta^{(t)}(\boldsymbol{r}) = \frac{ \boldsymbol{m}^{(t)}(\boldsymbol{r}) - \iint_{\mathbb{R}^2} \mathcal{K}(\boldsymbol{r} - \boldsymbol{r}') \cdot \boldsymbol{\delta}^{(t-1)}(\boldsymbol{r}') \, \mathrm{d} \boldsymbol{r}' }{ \sqrt{ \boldsymbol{v}^{(t)}(\boldsymbol{r}) } + \varepsilon }.
\end{equation}

Starting from the initial complex field $\boldsymbol{\Psi}^{(0)}(\boldsymbol{r})$, three such iterations accumulate to $\boldsymbol{\Psi}^{(3)}(\boldsymbol{r})$; enforcing the optical constraint then projects the result onto the feasible set $\mathbb{S}_z$, giving the final field:
\begin{equation}
\boldsymbol{\Psi}_{\text{final}}(\boldsymbol{r}) = 
\Pi_{\mathbb{S}_z} \left\{
\boldsymbol{\Psi}^{(0)}(\boldsymbol{r}) 
- \sum_{t=1}^{3} \gamma_t \cdot 
\Delta^{(t)}(\boldsymbol{r})
\right\},
\end{equation}
where $\gamma_t$ denotes the step size at iteration $t$, and $\Delta^{(t)}(\boldsymbol{r})$ is the update increment defined above.

\subsection{High-frequency Cross-layer Compensation Enhancement }

Indirect observation entails multibounce transport through intermediary surfaces;  along such energy paths the measured field undergoes severe attenuation, spectral smoothing, and detail depletion.  In the hidden-target region, multipath loss, non-ideal reflection, and ambient noise jointly degrade edge sharpness;  high-frequency geometric components, already faint, become difficult to resolve, which constrains structural-fidelity recovery.  To mitigate this bottleneck, Cross-layer Compensation forms two complementary streams with multi-receptive-field design and cross-level structural consistency: a frequency-selective branch that extracts residual high-frequency evidence via variably dilated band-pass kernels, avoiding over-smoothing inherent to purely spatial convolution; a semantic-anchoring branch that uses high-level global semantics as anchors to align low-level spatial detail, thereby suppressing noise-induced spurious boundaries.  The pipeline is illustrated in Figure \ref{fig:model}.

During propagation the optical signal experiences repeated reflections and frequency-dependent material attenuation;   cumulative loss grows with the reflection count $K$. Yet high-frequency components, less susceptible to absorption, may retain discernible traces in the final observation despite low energy. Mathematical modeling and frequency domain analysis appear in Appendix.

To explicitly target these components, we define a set of frequency-selective filters with spatial locality and adaptive receptive fields. Let $\mathcal{S}(\mathbf{r}, t)$ denote the input spatiotemporal signal, where $\mathbf{r} \in \mathbb{R}^2$ is spatial position and $t \in \mathbb{R}$ is time. Each branch $m$ uses a learnable band-pass kernel $\mathcal{G}_m$ with spatial dilation $\delta_m$ and scale $\sigma_m$, modulated by frequency response $\mathcal{W}_m$ (see Appendix for details).

For each filter branch with dilation $m \in \{1, 3, 5\}$, we define a Gaussian convolution operator that provides localized smoothing:
\begin{equation}
\mathcal{K}_\sigma[f](\mathbf{r}, t) := \int_{\mathbb{R}^2} f(\mathbf{r}', t) \cdot \exp\left(-\frac{\|\mathbf{r} - \mathbf{r}'\|^2}{2 \sigma^2}\right) \, d\mathbf{r}',
\end{equation}
where \( \mathbf{r}, \mathbf{r}' \in \mathbb{R}^2 \) are spatial coordinates, $\sigma \in \mathbb{R}^+$ controls the degree of spatial spread, regulating sensitivity to local variation. 

Filtered features are:
\begin{equation}
\mathcal{F}_m(\mathbf{r},t)\!=\!\iint_{\mathbb{R}^2 \times \mathbb{R}} \mathcal{G}_m(\mathbf{r} - \mathbf{r}', \omega'; \delta_m) \cdot \mathcal{Q}_m(\mathbf{r}', t, \omega') \, d\mathbf{r}' d\omega',
\end{equation}
where $\mathcal{Q}_m(\mathbf{r}', t, \omega') := \mathcal{S}(\mathbf{r}', t) \cdot \mathcal{A}_m(\mathbf{r}', \omega')$, \( \mathcal{S} \) is the input signal, \( \mathcal{A}_m(\mathbf{r}', \omega') = \sigma(\Theta_m \cdot \mathcal{F}_{\omega'}[\mathcal{S}(\mathbf{r}', t)]) \), \( \Theta_m \in \mathbb{R}^{C \times C} \) are learnable parameters, and \( \sigma(\cdot) \) is ReLU. Fused multi-branch features:
\begin{equation}
\mathcal{F}_{\text{fused}}(\mathbf{r}, t) = \sigma\left( \mathcal{K}_{\sigma_{\text{fuse}}} \left[ \sum_{m=1}^3 \mathcal{T}_m \cdot \mathcal{F}_m \right] \right),
\end{equation}
where \( \mathcal{T}_m \in \mathbb{R}^{C \times C} \) are fusion weights, \( \sigma_{\text{fuse}} \in \mathbb{R}^+ \).

The outputs of all branches are fused to form a unified frequency-enhanced representation:
\begin{equation}
\mathcal{F}_{\text{mfeb},k}(\mathbf{r}, t) = \mathcal{K}_{\sigma_k} \left[ \sum_{m=1}^{k} \mathcal{F}_m + \mathcal{H}_k + \mathcal{S} \right],
\end{equation}
where \( \mathcal{H}_k(\mathbf{r}, t) \) is the output of the \( k \)-th auxiliary convolution and \( \sigma_k \in \mathbb{R}^+ \).

We then apply channel-wise attention to reweight semantic channels by their spatial importance:
\begin{equation}
\begin{split}
\mathcal{C}(\mathbf{r}, c) &= \sigma\left( \mathcal{K}_{\sigma_c} \left[ \mathcal{F}_{\text{mfeb}} \cdot \Phi_c \right] \right), \\
\mathcal{F}_{\text{ca}}(\mathbf{r}, t) &= \sum_{c=1}^C \mathcal{C}(\mathbf{r}, c) \cdot \mathcal{F}_{\text{mfeb}}(\mathbf{r}, t, c),
\end{split}
\end{equation}
where \( \Phi_c \in \mathbb{R}^{C \times C} \), \( \sigma_c \in \mathbb{R}^+ \). Next, edge-aware filtering is applied by combining low-level and high-level paths:
\begin{equation}
\begin{split}
\mathcal{F}_{\text{edge,in}} &= \mathcal{K}_{\sigma_{\text{edge}}} \left[ \mathcal{F}_{\text{fused}} + \mathcal{F}_{\text{ca}} \right], \\
\mathcal{E}(\mathbf{r}, t) &= \mathcal{K}_{\sigma_{\text{ep}}} \left[ \Psi_{\text{edge}} \cdot \mathcal{F}_{\text{edge,in}} \right],
\end{split}
\end{equation}
where \( \Psi_{\text{edge}} \in \mathbb{R}^{C \times 1} \), \( \sigma_{\text{edge}}, \sigma_{\text{ep}} \in \mathbb{R}^+ \). 

We further include a spatial attention map to emphasize salient regions:
\begin{equation}
\mathcal{A}_{\text{sa}}(\mathbf{r}, t) = \sigma\left( \mathcal{K}_{\sigma_{\text{sa}}} \left[ \mathcal{F}_{\text{ca}} \cdot \Phi_{\text{sa}} \right] \right),
\end{equation}
where \( \Phi_{\text{sa}} \in \mathbb{R}^{C \times 1} \), \( \sigma_{\text{sa}} \in \mathbb{R}^+ \). 

We generate the saliency-enhanced features by combining all enhanced features and saliency cues:
\begin{equation}
\mathcal{F}_{\text{mea}}(\mathbf{r}, t) \!=\! \mathcal{K}_{\sigma_{\text{mea}}} \!\left[ \left( \mathcal{A}_{\text{sa}} \!+\! \sigma(\mathcal{E}) \right) \cdot \mathcal{F}_{\text{ca}} \!+\! \sum_{k=1}^3 \mathcal{F}_{\text{mfeb},k} \!+\! \mathcal{S} \right],
\end{equation}
where \( \sigma_{\text{mea}} \in \mathbb{R}^+ \). Final saliency convolution:
\begin{equation}
\mathcal{F}_{\text{out}}(\mathbf{r}, t) = \mathcal{K}_{\sigma_{\text{sal}}} \left[ \Psi_{\text{sal}} \cdot \mathcal{F}_{\text{mea}} \right],
\end{equation}
where \( \Psi_{\text{sal}} \in \mathbb{R}^{C \times C} \), \( \sigma_{\text{sal}} \in \mathbb{R}^+ \). 

We introduce cross-layer attention to inject high-level feedback into low-level features via a similarity-weighted aggregation. Cross-layer attention weight and response:
\begin{equation}
\alpha_l(\mathbf{r}, \mathbf{r}', \omega) = \sigma\left( \frac{ \int_{\mathbb{R}} \mathcal{Q}_{l-1}(\mathbf{r}, \omega') \cdot \mathcal{K}_l(\mathbf{r}', \omega') \, d\omega'}{ \sqrt{ \int_{\mathbb{R}^2} \int_{\mathbb{R}} \|\mathcal{K}_l(\mathbf{r}', \omega')\|^2 \, d\omega' \, d\mathbf{r}' } } \right),
\end{equation}
\begin{align}
\mathcal{A}_{l \to l-1}(\mathbf{r}, t) 
&= \int_{\mathbb{R}^2} \int_{\mathbb{R}} \alpha_l(\mathbf{r}, \mathbf{r}', \omega) \cdot \mathcal{V}_l(\mathbf{r}', \omega) \notag \\
&\quad \cdot \exp\left( -\frac{ \| \mathbf{r} - \mathbf{r}' \|^2 }{2 \sigma_{\text{attn}}^2} \right) \, d\omega \, d\mathbf{r}',
\end{align}
where \( \mathcal{Q}_{l-1}, \mathcal{K}_l, \mathcal{V}_l \in \mathbb{C} \), and \( \sigma_{\text{attn}} \in \mathbb{R}^+ \). 

Finally, the fused output incorporating both attention and saliency enhancement is:
\begin{equation}
\mathcal{F}_{\text{final}}(\mathbf{r}, t) = \mathcal{K}_{\sigma_{\text{final}}} \left[ \mathcal{A}_{l \to l-1} + \mathcal{F}_{\text{out}} \right],
\end{equation}
where \( \sigma_{\text{final}} \in \mathbb{R}^+ \).

The complete derivation is provided in Appendix. This framework compensates for low-frequency loss, leverages semantic priors, and enhances stability in low signal-to-noise conditions.

\begin{figure*}[ht]
    \centering
    \includegraphics[width=\linewidth]{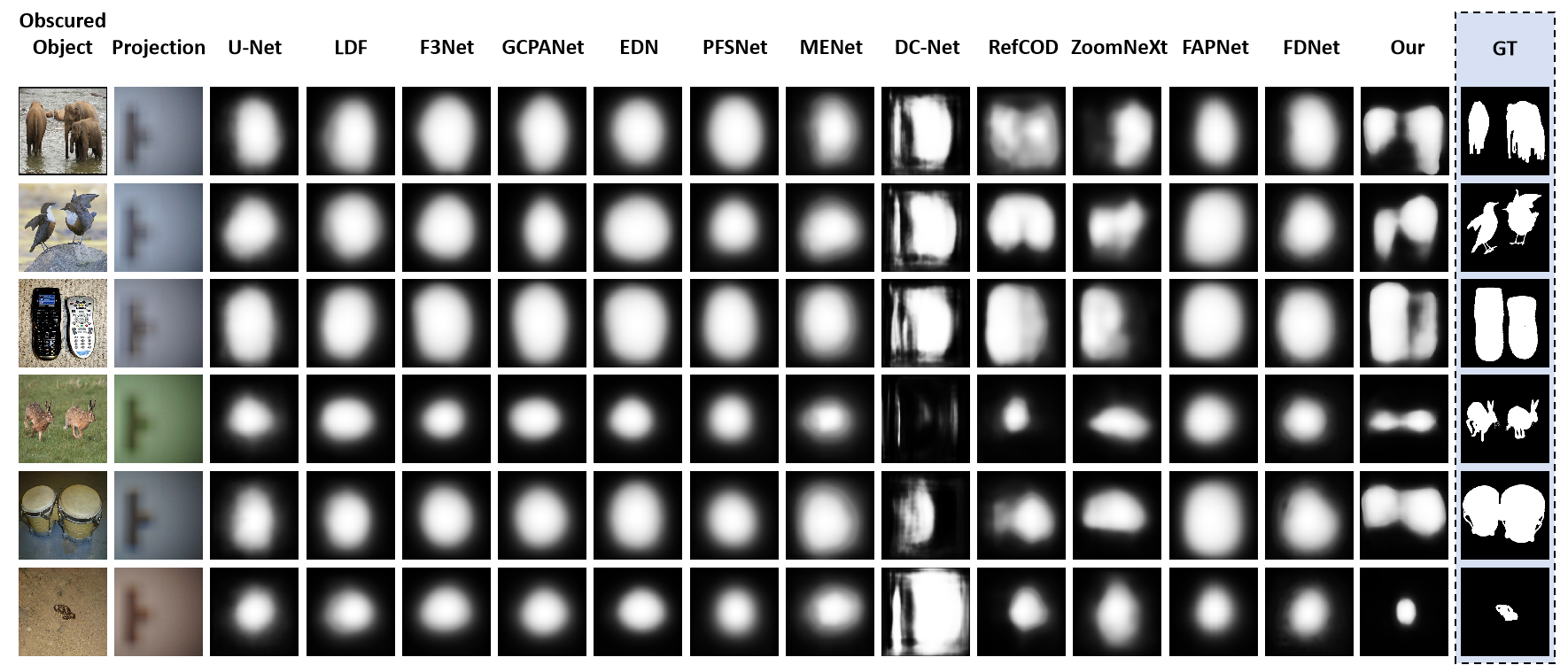}
    \caption{Visual comparison of detection results across representative methods on occluded-target datasets.}
    \label{fig:result}
\end{figure*}

\section{Experiments}

\begin{table*}[ht]
\centering
\small
{
\setlength{\tabcolsep}{1mm}
\begin{tabular}{c|cccccc|cccccc}
\toprule
\multirow{2}{*}{\textbf{Method}}  & \multicolumn{6}{c|}{\textbf{OOD-DUTS}} & \multicolumn{6}{c}{\textbf{OOD-HKU}} \\
\cmidrule(lr){2-7} \cmidrule(lr){8-13}
& Max-F & Mean-F & Fbw & MAE & SM & EM & Max-F & Mean-F & Fbw & MAE & SM & EM \\
\midrule
UNet \cite{ronneberger2015u} & 0.645 & 0.602  & 0.446 & 0.220  & 0.674 & 0.785 & 0.515 & 0.499  & 0.370  & 0.233 & 0.608 & 0.748 \\
LDF \cite{wei2020label} & 0.649 & 0.604  & 0.455 & 0.214 & 0.677 & 0.784 & 0.522 & 0.501  & 0.388 & 0.220  & 0.611 & 0.747 \\
F3Net \cite{wei2020f3net} & 0.646 & 0.603  & 0.451 & 0.215 & 0.676 & 0.783 & 0.514 & 0.498  & 0.379 & 0.227 & 0.608 & 0.744 \\
GCPANet \cite{chen2020global} & 0.646 & 0.600    & 0.456 & 0.211 & 0.677 & 0.781 & 0.520  & 0.501  & 0.383 & 0.221 & 0.612 & 0.751 \\
EDN \cite{wu2022edn} & 0.647 & 0.600    & 0.456 & 0.214 & 0.676 & 0.780  & 0.531 & 0.514  & 0.390  & 0.211 & 0.614 & 0.760 \\
PFSNet \cite{ma2021pyramidal} & 0.649 & 0.602  & 0.446 & 0.220  & 0.673 & 0.782 & 0.510 & 0.492  & 0.374 & 0.219 & 0.604 & 0.748 \\
MENet \cite{wang2023pixels} & 0.638 & 0.602  & 0.433 & 0.225 & 0.672 & 0.786 & 0.512 & 0.494  & 0.355 & 0.231 & 0.603 & 0.750 \\
DC-Net \cite{zhu2025dc} & 0.484 & 0.434  & 0.302 & 0.369 & 0.472 & 0.658 & 0.413 & 0.369  & 0.265 & 0.290  & 0.515 & 0.683 \\
RefCOD \cite{zhang2025referring} & 0.670  & 0.626  & 0.505 & 0.179 & 0.697 & 0.802 & 0.506 & 0.489  & 0.373 & 0.226 & 0.602 & 0.742 \\
ZoomNeXt \cite{pang2024zoomnext} & 0.672 & 0.629  & 0.495 & 0.186 & 0.700   & 0.802 & 0.532 & 0.510   & 0.389 & 0.224 & 0.615 & 0.746 \\
FAPNet \cite{zhou2022feature} & 0.648 & 0.604  & 0.459 & 0.207 & 0.680  & 0.785 & 0.520 & 0.504  & 0.387 & 0.213 & 0.610 & 0.756 \\
FDNet \cite{zhong2022detecting}  & 0.652 & 0.604  & 0.446 & 0.221 & 0.674 & 0.783 & 0.518 & 0.499  & 0.379 & 0.233 & 0.610  & 0.745 \\
\midrule
\textbf{Our}  & \textbf{0.692} & \textbf{0.654} & \textbf{0.550} & \textbf{0.154} & \textbf{0.708} & \textbf{0.823} & \textbf{0.564} & \textbf{0.545} & \textbf{0.439} & \textbf{0.170} & \textbf{0.610} & \textbf{0.785} \\
\bottomrule
\end{tabular}
}
\caption{Performance comparison on OOD-DUTS and OOD-HKU datasets under six evaluation metrics.}
\label{tab:baseline-part1}
\end{table*}

\begin{table*}[ht]
\centering
\small
{
\setlength{\tabcolsep}{1.1mm}
\begin{tabular}{c|c|cccccc|cccccc}
\toprule
\multirow{2}{*}{\textbf{Method}} & \multirow{2}{*}{\textbf{Publication}} & \multicolumn{6}{c|}{\textbf{OOD-ECSSD}} & \multicolumn{6}{c}{\textbf{OOD-DUT-OMRON}} \\
\cmidrule(lr){3-8} \cmidrule(lr){9-14}
& & Max-F & Mean-F & Fbw & MAE & SM & EM & Max-F & Mean-F & Fbw & MAE & SM & EM \\
\midrule
UNet                    & MICCAI 2015                  & 0.604 & 0.557  & 0.384 & 0.258 & 0.628 & 0.721 & 0.436 & 0.402  & 0.292 & 0.203 & 0.583 & 0.671 \\
LDF                     & CVPR 2020                    & 0.610 & 0.565  & 0.401 & 0.240 & 0.636 & 0.726 & 0.447 & 0.407  & 0.312 & 0.191 & 0.588 & 0.673 \\
F3Net                   & AAAI 2020                    & 0.608 & 0.566  & 0.406 & 0.237 & 0.641 & 0.726 & 0.435 & 0.400    & 0.303 & 0.195 & 0.583 & 0.67  \\
GCPANet                 & AAAI 2020                    & 0.600 & 0.56   & 0.394 & 0.243 & 0.635 & 0.723 & 0.436 & 0.399  & 0.305 & 0.195 & 0.584 & 0.670 \\
EDN                     & TIP 2022                     & 0.618 & 0.562  & 0.435 & 0.223 & 0.649 & 0.723 & 0.464 & 0.422  & 0.326 & 0.193 & 0.597 & 0.681 \\
PFSNet                  & AAAI 2021                    & 0.612 & 0.565  & 0.427 & 0.232 & 0.644 & 0.726 & 0.447 & 0.411  & 0.314 & 0.192 & 0.592 & 0.677 \\
MENet                   & CVPR 2023                    & 0.605 & 0.557  & 0.381 & 0.241 & 0.622 & 0.724 & 0.442 & 0.404  & 0.297 & 0.207 & 0.585 & 0.676 \\
DC-Net                  & Pattern Recognit. 2024     & 0.466 & 0.401  & 0.302 & 0.402 & 0.449 & 0.595 & 0.346 & 0.312  & 0.218 & 0.279 & 0.508 & 0.628 \\
RefCOD                  & TPAMI 2025                   & 0.615 & 0.569  & 0.435 & 0.226 & 0.646 & 0.727 & 0.421 & 0.387  & 0.288 & 0.220  & 0.571 & 0.656 \\
ZoomNeXt                & PAMI2024                     & 0.605 & 0.546  & 0.415 & 0.245 & 0.630  & 0.707 & 0.449 & 0.408  & 0.310  & 0.195 & 0.590  & 0.672 \\
FAPNet                  & TIP2022                      & 0.603 & 0.559  & 0.400 & 0.242 & 0.638 & 0.723 & 0.455 & 0.411  & 0.327 & 0.194 & 0.593 & 0.673 \\
FDNet                   & CVPR2022                     & 0.603 & 0.554  & 0.384 & 0.255 & 0.627 & 0.720 & 0.441 & 0.402  & 0.301 & 0.200 & 0.585 & 0.667 \\
\midrule
\textbf{Our} &\textbf{-}  & \textbf{0.666} & \textbf{0.634} & \textbf{0.533} & \textbf{0.159} & \textbf{0.667} & \textbf{0.787} & \textbf{0.471} & \textbf{0.433} & \textbf{0.341} & \textbf{0.169} & \textbf{0.603} & \textbf{0.695} \\
\bottomrule
\end{tabular}
}
\caption{Performance comparison on OOD-ECSSD and OOD-DUT-OMRON datasets under six evaluation metrics.}
\label{tab:baseline-part2}
\end{table*}

We evaluate WavePCNet on four physically constructed datasets: OOD-DUTS, OOD-HKU, OOD-ECSSD, and OOD-DUT-OMRON, which are derived from the corresponding public datasets DUTS\cite{Wang_2017_CVPR}, HKU-IS\cite{Li_2015_CVPR}, ECSSD\cite{Yan_2013_CVPR}, and DUT-OMRON\cite{Yang_2013_CVPR} by photographing printed scenes at our experimental site. The evaluation is conducted using six standard metrics: Maximum F-measure (Max-F), Mean F-measure (Mean-F), Weighted F-measure (Fbw), Mean Absolute Error (MAE), Structure-measure (S-measure), and Enhanced-alignment Measure (E-measure). 

WavePCNet achieves state-of-the-art performance on both the OOD-DUTS and OOD-HKU datasets, as shown in Table \ref{tab:baseline-part1}. Notably, on OOD-DUTS, it achieves a Max-F score of 0.692, which is a 4.0\% absolute improvement over the best existing method (RefCOD, 0.670), and reduces MAE to 0.154, the lowest among all methods, demonstrating superior boundary precision and noise resilience. Similarly, on OOD-HKU, WavePCNet yields the highest Mean-F (0.545) and Fbw (0.439), and the lowest MAE (0.170), outperforming the second-best method (ZoomNeXt) by 1.6\% in Mean-F and 2.3\% in Fbw. Table \ref{tab:baseline-part2} shows the results on the OOD-ECSSD and OOD-DUT-OMRON datasets. On OOD-ECSSD, WavePCNet achieves the highest scores in all metrics, including Max-F (0.666) and Fbw (0.533), and achieves the lowest MAE (0.159), an absolute improvement of 5.1\% in Max-F and reduction of 0.081 in MAE, compared to the next best method (RefCOD). On OOD-DUT-OMRON, WavePCNet maintains top performance with a Max-F of 0.471, outperforming the best prior method (ZoomNeXt, 0.456) by 1.5\%, and achieves the lowest MAE (0.169) and highest SM (0.603). Furthermore, Figure \ref{fig:result} shows the comparison of the detection results. Compared to prior approaches, WavePCNet consistently produces results more accurate and closer to the ground truth, with sharper boundaries, clearer object shapes, and fewer artifacts. While methods like U-Net, F3Net, and GCPANet often generate blurred or incomplete outputs, and others like DC-Net and MENet fail under heavy occlusion, WavePCNet preserves fine structure and semantic integrity even in challenging scenes.

Overall, real-valued, locality-biased convolutions in existing architectures, constrained by their value-domain modeling, respond poorly to phase aliasing and non-specular diffusion, hence information loss in low-SNR, multiply scattered fields. By contrast, TriWCP in WavePCNet introduces complex-domain propagation with a three-stage optimisation and low-rank spectral constraint; update trajectories remain confined to a physics-consistent propagating subspace, expected to improve recovery of signals from occluded regions. Additionally, Cross-layer Compensation, coupling multi-scale frequency-selective filtering with cross-layer attention guidance, extracts residual high-frequency responses under strong scattering backgrounds and aligns structural semantics across levels, thereby providing finer edge resolution and stronger context awareness for obscured objects.

\begin{table}[htbp]
\centering
\setlength{\tabcolsep}{1.2pt}
\resizebox{\linewidth}{!}{%
\begin{tabular}{cc|cccccc}
\toprule
TriWCP & Compen & Max-F & Mean-F & Fbw  & MAE   & SM   & EM   \\
\midrule
\ding{55} & \ding{51} & 0.610 & 0.557 & 0.424 & 0.237 & 0.636 & 0.715 \\
\ding{51} & \ding{55} & 0.477 & 0.429 & 0.284 & 0.273 & 0.535 & 0.682 \\
\midrule
MobileNet       & \ding{51} & 0.598 & 0.566 & 0.454 & 0.186 & 0.603 & 0.745 \\
VGG             & \ding{51} & 0.607 & 0.569 & 0.482 & 0.186 & 0.618 & 0.746 \\
ShuffleNetV2    & \ding{51} & 0.592 & 0.555 & 0.452 & 0.196 & 0.620 & 0.743 \\
EfficientNet-B0 & \ding{51} & 0.587 & 0.549 & 0.452 & 0.195 & 0.606 & 0.734 \\
\midrule
\ding{51} & UNet        & 0.622 & 0.583 & 0.475 & 0.190 & 0.641 & 0.749 \\
\ding{51} & FPN         & 0.616 & 0.575 & 0.453 & 0.195 & 0.635 & 0.747 \\
\ding{51} & HRNet       & 0.634 & 0.592 & 0.480 & 0.184 & 0.654 & 0.757 \\
\ding{51} & Transformer & 0.608 & 0.570 & 0.431 & 0.225 & 0.639 & 0.734 \\
\midrule
\ding{51} & \ding{51} & \textbf{0.666} & \textbf{0.634} & \textbf{0.533} & \textbf{0.159} & \textbf{0.667} & \textbf{0.787} \\
\bottomrule
\end{tabular}%
}
\caption{Ablation configurations evaluating the isolated and joint contributions of TriWCP and Cross-layer Compensation (Compen) under indirect, multiply-scattered conditions.}
\label{tab:Ablation_single_column}
\end{table}

\subsection{Ablation Studies}
To verify the independent efficacy, and the joint contribution of TriWCP and Cross-layer Compensation under indirect, multiply scattered propagation, three experiment groups are configured, the results are shown in Table \ref{tab:Ablation_single_column}.

\textbf{Two ablations (removed TriWCP or Compensation respectively)}: the full model achieves the best performance, with a Max-F of 0.666, Mean-F of 0.634, and the lowest MAE of 0.159. Compared to using only Compensation, the full model improves Max-F by 5.6\% and reduces MAE by 0.078, while compared to using only TriWCP, it shows an even larger gain of 18.9\% in Max-F and 0.114 reduction in MAE, highlighting the importance of both modules.

\textbf{Compen with real-valued convolutional backbones(including MobileNet and VGG)}: 
VGG achieves the best results with a Max-F of 0.607 and Mean-F of 0.569, while EfficientNet-B0 performs the worst, with a lower Max-F (0.587) and the highest MAE (0.195). Although VGG slightly outperforms others, all backbone-only variants significantly underperform compared to the full WavePCNet, achieving up to a 9.1\% increase in Max-F (over EfficientNet-B0) and a 0.036 reduction in MAE (over MobileNet and VGG). These results highlight the performance of TriWCP.

\textbf{ TriWCP with classical fusion frameworks (e.g., UNet and FPN)}: HRNet performs best (Max-F: 0.634, MAE: 0.184), compared to the full WavePCNet, which achieves a Max-F of 0.666 and MAE of 0.159, the improvement reaches up to a 5.8\% increase in Max-F (over Transformer) and 0.066 reduction in MAE. Even over the strongest baseline (HRNet), WavePCNet still improves Max-F by 3.2\% and reduces MAE by 0.025. 

Overall, the comparative analyses across various backbone configurations demonstrate the critical importance of TriWCP and Cross-layer Compensation beyond simply selecting a strong backbone, and the synergy between physics-guided coherent propagation and frequency-aware structural enhancement is key to robust obscured object detection under passive, non-line-of-sight conditions.

\subsection{Complexity and Performance Trade-off}

\begin{table}[t]
\centering
\setlength{\tabcolsep}{2.2pt}
\resizebox{\linewidth}{!}{%
\begin{tabular}{c|ccc}
\toprule
Method & Params (M) $\downarrow$ & FPS $\uparrow$ & MAE $\downarrow$ \\
\midrule
UNet        & 23.95 & 186.270 & 0.258 \\
LDF             & 25.15 & 116.433 & 0.240 \\
F3Net          & 25.54 & 107.250 & 0.237 \\
GCPANet       & 67.06 &  94.307 & 0.243 \\
EDN                & 42.90 &  65.445 & 0.223 \\
PFSNet       & 31.18 &  93.118 & 0.232 \\
MENet         & 27.83 &  26.986 & 0.241 \\
DC-Net             & 284.11&  30.346 & 0.402 \\
RefCOD    & 24.98 & 109.188 & 0.226 \\
ZoomNeXt    & 28.46 &  38.646 & 0.245 \\
FAPNet       & 24.84 &  91.110 & 0.242 \\
FDNet     & 24.10 & 178.444 & 0.255 \\
\midrule
\textbf{WavePCNet (Ours)}           & \textbf{26.05} & \textbf{39.428} & \textbf{0.159} \\
\bottomrule
\end{tabular}%
}
\caption{Comparison of computational/storage complexity (Params), runtime throughput (FPS), and reconstruction fidelity (MAE).}
\label{tab:complexity_fps_mae}
\end{table}

As shown in Table~\ref{tab:complexity_fps_mae}, WavePCNet achieves a favorable complexity–speed–accuracy trade-off: it uses 26.05M parameters—comparable to UNet (23.95M) and F3Net (25.54M), and far smaller than GCPANet (67.06M) or DC-Net (284.11M). It runs at 39.428 FPS, meeting real-time needs (slower than UNet/FDNet but faster than MENet, 26.986 FPS, and comparable to SSN, 37.821 FPS). Most importantly, it delivers the best accuracy with an MAE of 0.159, outperforming EDN (0.223) and RefCOD (0.226), attributable to the proposed physics-driven TriWCP and cross-layer frequency compensation.

\section{Conclusion}

This study pioneers the definition and validation of the passive post-corner OOD task. Operating without active illumination and relying solely on faint ambient-scattered returns, WavePCNet enforces coherent propagation consistency in the complex domain, augmented by cross-layer frequency compensation, to jointly localize and segment targets hidden beyond the line of sight. The TriWCP module constrains the optimization process to a physically meaningful low-rank subspace, mitigating instability caused by wavefront distortion. Meanwhile, the Cross-layer Compensation module amplifies residual high-frequency features and restores boundary fidelity through semantics-guided structural alignment. Comprehensive evaluations and ablation studies across four physically collected datasets validate the robustness and complementarity of the proposed components, highlighting their effectiveness in tackling the unique challenges of passive OOD.


\bibliography{aaai2026}

\end{document}